%% file: wider-2018-summary.tex
\documentclass[10pt,journal,compsoc]{IEEEtran}
\ifCLASSOPTIONcompsoc
  \usepackage[nocompress]{cite}
\else
  \usepackage{cite}
\fi
\ifCLASSINFOpdf
\else
\fi
\usepackage{amsmath}
\usepackage{url}

\usepackage{xcolor}

\usepackage{graphicx}
\graphicspath{{figures}}

\begin{document}
%
\title{WIDER Face and Pedestrian Challenge 2018 \\ Methods and Results}
%
%
%
%

\author{
	Chen Change Loy, Dahua Lin, Wanli Ouyang,
	Yuanjun Xiong, Shuo Yang, Qingqiu Huang,
	Dongzhan Zhou, Wei Xia, Quanquan Li, Ping Luo, Junjie Yan,
	Jianfeng Wang, Zuoxin Li, Ye Yuan, Boxun Li, Shuai Shao, Gang Yu,
	Fangyun Wei, Xiang Ming, Dong Chen,
	Shifeng Zhang, Cheng Chi, Zhen Lei, Stan Z. Li,
	Hongkai Zhang, Bingpeng Ma, Hong Chang, Shiguang Shan, Xilin Chen,
	Wu Liu, Boyan Zhou, Huaxiong Li, Peng Cheng, Tao Mei,
	Artem Kukharenko, Artem Vasenin, Nikolay Sergievskiy,
	Hua Yang, Liangqi Li, Qiling Xu, Yuan Hong, Lin Chen, Mingjun Sun,
	Yirong Mao, Shiying Luo, Yongjun Li, Ruiping Wang,
	Qiaokang Xie, Ziyang Wu, Lei Lu, Yiheng Liu, Wengang Zhou
\IEEEcompsocitemizethanks{
\IEEEcompsocthanksitem C. C. Loy are with Nanyang Technological University, Singapore.
\IEEEcompsocthanksitem Q. Huang, P. Luo, and D. Lin are with The Chinese University of Hong Kong.
\IEEEcompsocthanksitem S. Yang, Y. Xiong, and  W. Xia are with Amazon Web Services.
\IEEEcompsocthanksitem D. Zhou and W. Ouyang are with University of Sydney.
\IEEEcompsocthanksitem Q. Li, J. Yan are with SenseTime.
\IEEEcompsocthanksitem J. Wang, Z. Li, Y. Yuan, B. Li, S. Shao, and G. Yu are with Megvii.
\IEEEcompsocthanksitem F. Wei, X. Ming, and D. Chen are with Microsoft Research Asia.
\IEEEcompsocthanksitem S. Zhang, C. Chi, Z. Lei, and S. Z. Li are with University of Chinese Academy of Sciences.
\IEEEcompsocthanksitem H. Zhang, B. Ma, H. Chang, Y. Mao, S. Luo, Y. Li, R. Wang, S. Shan, and X. Chen are with Institute of Computing Technology, Chinese Academy of Sciences.
\IEEEcompsocthanksitem W. Liu, B. Zhou, H. Li, P. Cheng, and T. Mei are with Computer Vision and Multimedia Lab, JD AI Research.
\IEEEcompsocthanksitem A. Kukharenko, A. Vasenin, and N. Sergievski are with NtechLab.
\IEEEcompsocthanksitem H. Yang, L. Li, Q. Xu, Y. Hong, L. Chen, and M. Sun are with Shanghai Jiao Tong University.
\IEEEcompsocthanksitem Q. Xie, Z. Wu, L. Lu, Y. Liu, and W. Zhou are with University of Science and Technology of China and iFLYTEK.
}
}

%
%

\markboth{}%
{Shell \MakeLowercase{\textit{et al.}}: The 2018 WIDER \\ Face and Pedestrian Challenge}
%



\IEEEtitleabstractindextext{%
\begin{abstract}
This paper presents a review of the 2018 WIDER Challenge on Face and Pedestrian.
The challenge focuses on the problem of precise localization of human faces and bodies, and accurate association of identities. It comprises of three tracks: 
(i)
WIDER Face which aims at soliciting new approaches to advance the state-of-the-art in face detection, 
(ii)
WIDER Pedestrian which aims to find effective and efficient approaches to address the problem of pedestrian detection in unconstrained environments,
and
(iii) WIDER Person Search which presents an exciting challenge of searching persons across 192 movies.
In total, 73 teams made valid submissions to the challenge tracks. 
We summarize the winning solutions for all three tracks.
and present discussions on open problems and potential research directions in these topics. 
\end{abstract}

}

\maketitle

\IEEEdisplaynontitleabstractindextext

%

\IEEEraisesectionheading{\section{Introduction}\label{sec:introduction}}

\vspace{-0.1cm}

%
%
%
%
Faces and persons are among the most researched subjects in computer vision.
The past years have seen many exciting progresses in analyzing human and human faces, during which public benchmarks and challenges have been an important driving force.
Inspired by the success of the ImageNet Challenge series~\cite{olga2015imagenet} and COCO Challenges~\cite{lin2014mscoco}, the 2018 WIDER Face and Pedestrian Challenge Workshop is organized on October 8, 2018 in conjunction with ECCV 2018 in Munich, Germany. 
This challenge comprises of three tasks and evaluation tracks: 1) face detection, 2) pedestrian detection, and 3) person search.
In the remaining sections we will provide summaries of the winning solutions and provide analysis to strength and limitation of the submissions. By the analysis we hope to take a closer look at the current state of the fields related to the challenge tasks.

\subsection{Challenge Summary}
In the 2018 WIDER Face and Pedestrian Challenge, three challenge tasks are established with their benchmark dataset provided to the participants. 
The challenge tracks are hosted separately on the CodaLab website\footnote{\url{http://codalab.org/}}. Participants are requested to upload algorithm output to the public evaluation server for each track. 
Each challenge track is divided into a \emph{validation} phase and a \emph{final test} phase. In the validation phase the participants are provided with a set of validation test and the groundtruth annotations.
The participants are allowed to upload submissions to the public evaluation server for validating their submissions.
In the final test phase, the participants are provided with another set of testing data without annotations. 
The models' performance metrics in the phase are used to determine the challenge winners.
The validation phase started on May 10, 2018\footnote{We will keep the evaluation server available before further notice.}. 
The final test phase started on June 18, 2018 and ended on July 18, 2018.
In total 73 teams made valid submissions to the challenge tracks. 
Three winning teams are determined for each track based on the evaluation metrics for the final test phase.

\input{sections/face_detection}
\input{sections/pedestrian_detection}
\input{sections/person_search}

\section{Conclusion}
In the three challenge tracks discussed above, different aspects of visual recognition of face and pedestrian are examined. 
We are glad to observe the winning submissions achieve promising performances on the challenge tracks. In the face detection and pedestrian detection tracks, we see a wide adoption of general object detection methods. Various training and inference techniques are proposed. 
For the person search task, a solid baseline based on the two-stage architecture is widely used. 
In the next challenge, we hope to provide a larger-scale training and evaluation data. We also hope to see new approaches being developed for each specific area of the challenge tracks.

\vspace{0.1cm}
\noindent\textbf{Acknowledgements}. We thank the WIDER Challenge 2018 sponsors:
SenseTime Group Ltd. and Amazon Rekognition.


%





\ifCLASSOPTIONcaptionsoff
  \newpage
\fi



%
\bibliographystyle{IEEEtran}
\bibliography{wider-2018-summary}

%





\end{document}

%% file: sections/face_detection.tex
\section{Face Detection Track}
\label{sec: face-detection}

\subsection{Task and Dataset}
Face detection is an important and long-standing problem in computer vision. Given an arbitrary image, the goal of face detection is to determine whether or not there are any faces in the image and, if present, return the image location and extent of each face ~\cite{yang2002detecting}. While this appears as an effortless task for a human, it is a very difficult task for computers. The challenges associated with face detection can be attributed to variations in pose, scale, facial expression, occlusion, lighting condition, etc.

\begin{figure}[t]
    \centering
    \includegraphics[width=\linewidth]{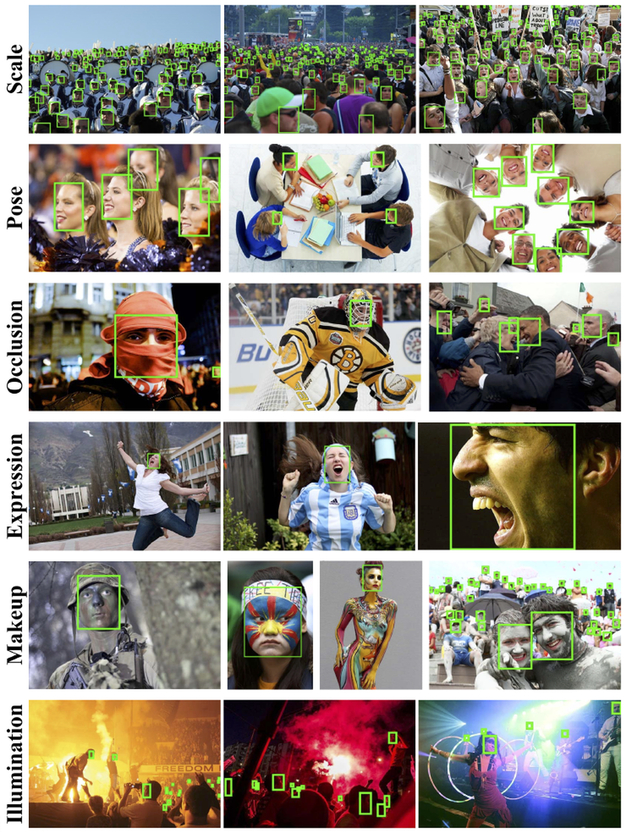}
    \caption{ We show example images (cropped) and annotations from WIDER FACE dataset.}
    \label{fig:wider_face_example}
\end{figure}

In this challenge, we choose WIDER FACE~\cite{yang2016wider} dataset as the benchmark. WIDER FACE dataset is currently the largest face detection dataset. It contains $32, 203$ images and $393, 703$ annotated faces with a high degree of variability in scale, pose and occlusion as depicted in Fig.~\ref{fig:wider_face_example}. WIDER FACE dataset is organized based on $60$ event classes. For each event class, we randomly select 40\%/10\%/50\% data as training, validation and testing sets.

\subsection{Evaluation Metric}

This section describes the detection evaluation metrics used by WIDER Face Challenge. The average precision is used for characterizing the performance of a face detector on WIDER FACE. Similar with COCO~\cite{lin2014microsoft} Challenge, AP is averaged over multiple Intersection over Union (IoU) values. Specifically, we use 10 IoU thresholds of .50:.05:.95. Averaging over IoUs rewards detectors with better localization.
    
Note that, there are a large number of small faces in WIDER FACE dataset. We mark faces with a height not fewer than 10 pixels as valid ground truth and the others as difficult samples. Similar to the evaluation of PASCAL VOC~\cite{everingham2010pascal}, difficult samples are allowed to be hit once without punishment. Different from the WIDER FACE dataset that contains three subsets for evaluation. We only evaluate WIDER FACE hard set for this challenge.

\subsection{Results of the Challenge}

The results of the top5 teams are shown in Tab.~\ref{tab:face-detection-results}.

\begin{table}[!h]
\renewcommand{\arraystretch}{1.3}
\caption{Results of the Top3 Teams in Face Detection Track}
\label{tab:face-detection-results}
\centering
\begin{tabular}{c|c}
    \hline
    Team & mAP (\%) \\
    \hline
    \hline    
    Megvii & \textbf{55.82} \\
    MSRA & 53.32 \\
    CASFD & 50.30\\
    ayantian & 50.24 \\
    yttrium & 49.78 \\
    \hline
\end{tabular}
\end{table}

\subsection{Solution of First Place}

The champion designs a single stage detector which applied multiple techniques published in recent years. The final results are generated by aggregating the predictions from multiple face detectors.

\noindent\textbf{Framework}.
The winning team proposed a single stage detector with the network structure based on RetinaNet~\cite{lin2018focal} and FAN~\cite{wang2017fan}. Similar to recent works~\cite{wang2017fan,wang2018sface}, the anchors are carefully designed based on the statistic of the training set. Data augmentations are applied. In order to boost the localization performance, a cascade structure is optimized in an end-to-end fashion.  

\noindent\textbf{Implementation Details}.
(1)  A square patch from the original image is first randomly cropped and then resized to the scale of $600\times600$ with the ground truth inside the patch used for training. Random horizontal flip and color jitter are also applied.
(2) Deformable Convolution~\cite{dai2017deformable} and IoU Loss~\cite{yu2016unitbox} are used to improve the detection performance.
(3) During testing, for each model, images with multi-scales and flips are used. The short edge of image is set to be $600$, $1,200$, $1,800$ and $2,400$ in the multi-scale test phase. Five models are selected for the ensemble to generate the final results.

\subsection{Solution of Second Place}

\begin{figure}[h]
    \centering
    \includegraphics[width=\linewidth]{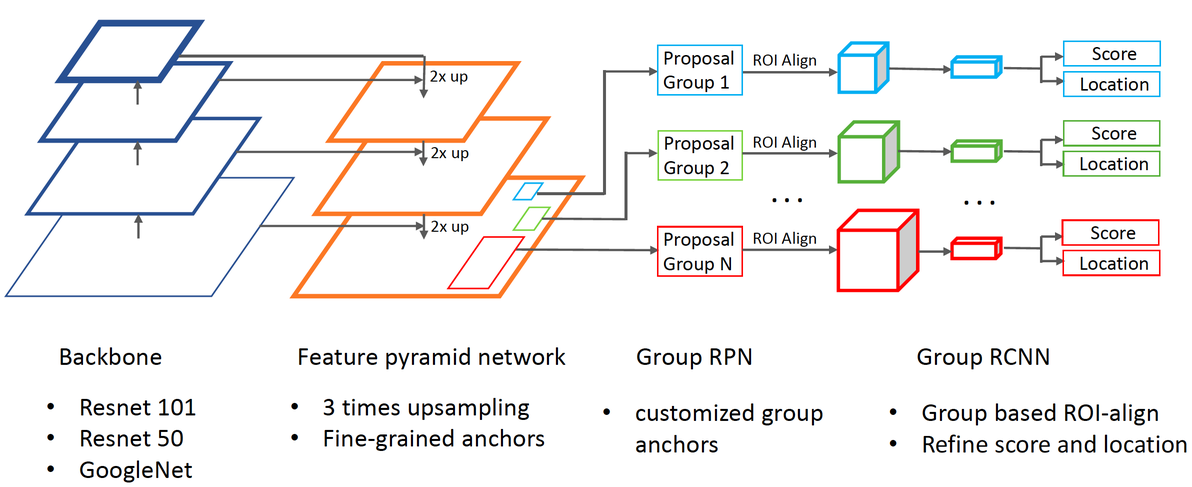}
    \caption{Framework of the 2nd place solution in Face Detection.}
    \label{fig:face_det_2ndt}
\end{figure}

The team in the second place proposed a two-stage face detector following Faster R-CNN~\cite{ren2017faster} and FPN~\cite{lin2017feature} framework. The overall framework is depicted in Fig.~\ref{fig:face_det_2ndt}.

\noindent\textbf{Framework}.
The team follows a two-stage Faster R-CNN~\cite{ren2017faster} framework. In the first stage, FPN-like architecture~\cite{lin2017feature} is adopted to provide rich context information for face detection, especially for small faces. Different from FPN~\cite{lin2017feature} which extracts different features on feature pyramid for different scales, the team only uses the feature from the finest level. To handle large-scale variance, the team divides proposals generated from the first step into several groups according to their scale and evenly sampled in each group. In the second stage, they use ROIAlign~\cite{he2017mask} to extract features for each proposal. A small network is used for each group to get the detection score.

\noindent\textbf{Implementation Details}.
(1)  In the first stage, anchors with different scales ${16; 32; 64; 128}$ are used in RPN. For each proposal group, the team randomly samples $2,048$ proposals used for the second stage. The feature of each proposal is then extracted using ROIAlign following the setting used in ~\cite{he2017mask} to set the pooling size $7 \times 7$.
(2) Two data augmentation strategies are applied. An image pyramid with the scale of $[0.25, 0.5, 0.75, 1, 1.25, 1.5, 1.75, 2]$ are first generated. During training, they randomly select an image from a pyramid and randomly crop a patch from a selected image to ensure that each side of the patch does not exceed 900 pixels. After that, a random horizontal flip is applied.
(3) During testing, the team scales both raw and horizontal flip images by a factor of $[0.125, 0.25, 0.5, 0.75, 1, 1.25, 1.5, 1.75, 2]$.

\subsection{Solution of Third Place}

\begin{figure}[h]
    \centering
    \includegraphics[width=\linewidth]{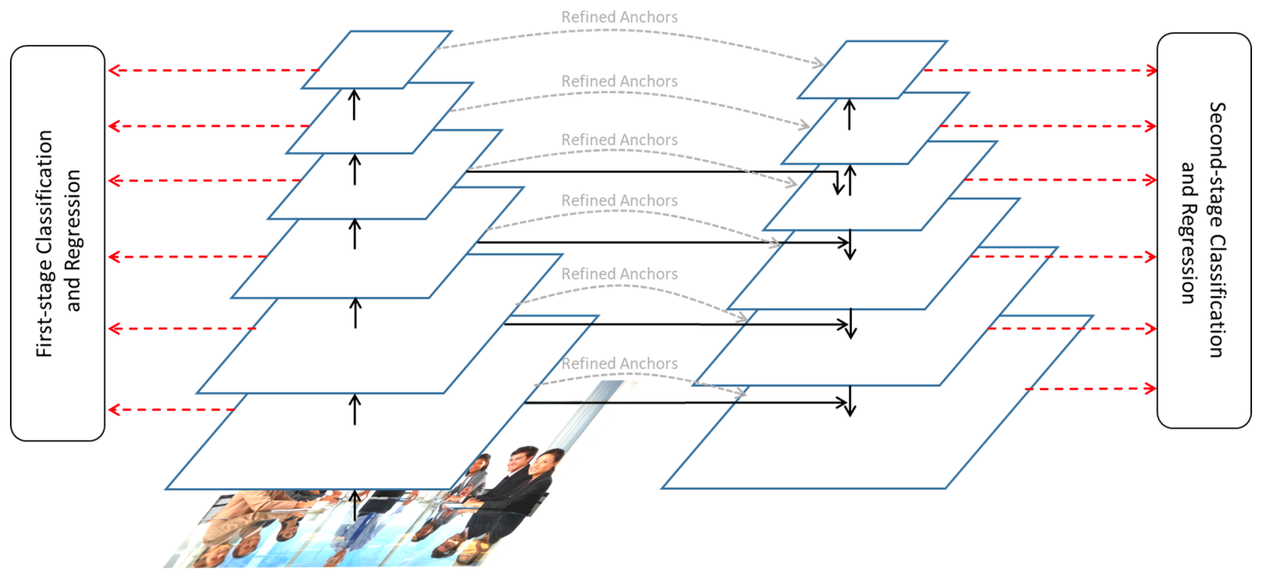}
    \caption{Framework of the 3rd place solution in Face Detection.}
    \label{fig:face_det_3rdt}
\end{figure}

The third place team proposed a two-stage face detection framework. The overall framework is depicted in Fig.~\ref{fig:face_det_3rdt}.

\noindent\textbf{Framework}.
The team in the third place follows RetinaNet~\cite{lin2018focal} and RefineDet~\cite{zhang2018refine} to design the backbone structure. The team uses two-stage classification and regression to improve the accuracy of classification and bounding boxes regression. To achieve a good recall rate, the anchors are carefully designed and focal loss and two-level classification are applied to reduce easy samples.

\noindent\textbf{Implementation Details}.
(1) Multi-scale augmentation is used in the training stage.
(2) To allow more faces to fall within the best detection range of the model, a multi-scale testing strategy is used in the test phase.

\subsection{Discussion}

The WIDER FACE dataset has a high degree of variability in scale and it contains a large number of tiny faces. In order to improve the recall rate without dramatically increasing the number of proposals, anchor boxes are carefully designed based on the statistics computed from the training set. Most the teams follow the recent advances in object detection~\cite{lin2017feature,dai2017deformable,lin2018focal,he2017mask} and face detection~\cite{zhang2018refine, wang2017fan,wang2018sface} to design their backbone network and powerful data augmentations are applied, e.g., multi-scale training. Since our evaluation metric emphasizes on the bounding boxes regression accuracy, cascade structures and multi-stage regression/classification are widely used. The winning teams have achieved remarkable face detection performance. However, novel ideas, especially those with computational cost considered, are rarely seen. We hope in the next round of the challenge we will be able to see more novel and efficient methods for solving the face detection problem. 

%% file: sections/pedestrian_detection.tex

\section{Pedestrian Detection Track}
\label{sec: pedestrian-detection}

\subsection{Task and Dataset}

The main goal of the Pedestrian Detection track is to address the problem of detecting pedestrians and cyclists in unconstrained environments. The dataset mainly considers two scenarios, surveillance and car-driving. To achieve satisfactory performance, participants need to design methods which can deal with the two scenarios at the same time. 

\begin{figure}[h]
    \centering
    \includegraphics[width=\linewidth]{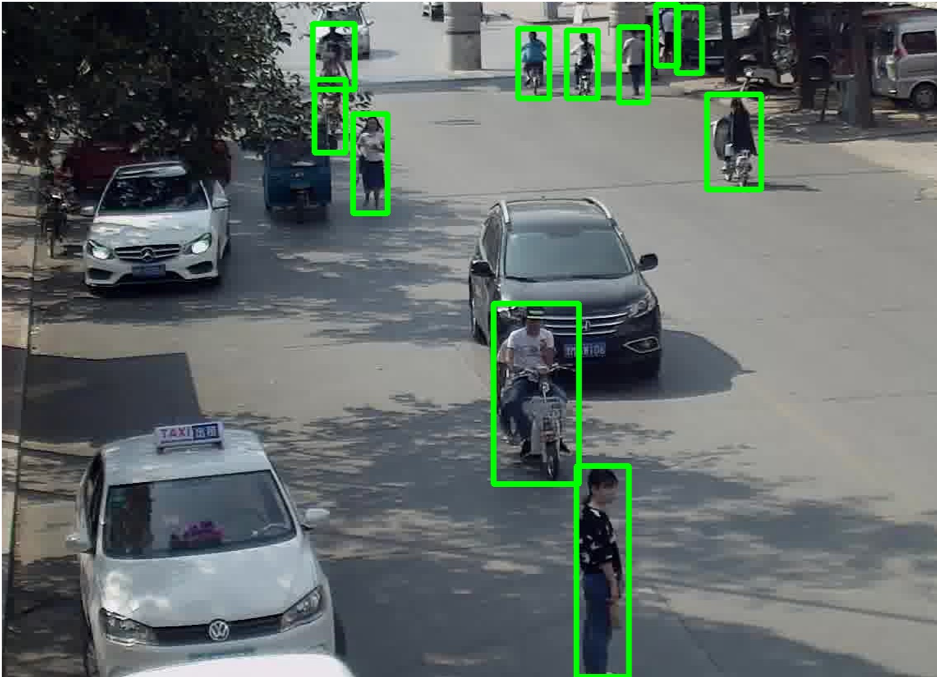}
    \caption{An Example of WIDER Pedestrian dataset.}
    \label{fig:pedestrian_sample}
\end{figure}

The dataset of WIDER Pedestrian Track (see Fig.~\ref{fig:pedestrian_sample}) contains a total of 20,000 images, half of which come from surveillance cameras and the other half from cameras located on driving vehicles through regular traffic in urban environments. There are 11,500 images in training, 5,000 in validation and 3,500 in the test. The total number of pedestrians and cyclists in the training and validation set are 46513 and 19696, respectively. We provide two categories in the training and validation sets, walking pedestrians as label 1 and cyclists as label 2. In the test stage, we do not distinguish the two categories. That is, the participants only need to submit the confidence and bounding boxes of all the pedestrians and cyclists detected and do not need to provide the categories.

Compared with previous competitions, WIDER Pedestrian brings more challenges in various aspects. The first challenge is the variety of data. The images from the two scenarios are very different in camera angle, object scale and illumination, so participants must propose robust and versatile methods. Moreover, many images are captured from a night scene, making the detection more difficult. Other challenges arise from the density of targets, smaller pedestrian scales and occlusions. All these factors place higher demands on the participating solutions.

\subsection{Evaluation Metric}

WIDER Pedestrian uses mean average precision (mAP) as the evaluation metric. The metric is the same as COCO Detection Task~\cite{lin2014mscoco}. The winners are determined by the Average AP over the 10 Intersection over Union (IoU) thresholds:.50:.05:.95. 
We delete submitted objects whose overlap ratio with the `ignore' region is more than $50\%$ in the evaluation stage. Meanwhile, the ground-truth objects which are in the same conditions with the `ignore' region will also be removed. In other words, we only use the objects in the non-ignoring parts to compute the final Average AP. 

\subsection{Results of the Challenge}

The results of the top5 teams are shown in Tab.~\ref{tab:pedestrian-detection}.

\begin{table}[!h]
\renewcommand{\arraystretch}{1.3}
\caption{Results of the Top5 Teams in Pedestrian Detection Track}
\label{tab:pedestrian-detection}
\centering
\begin{tabular}{c|c}
    \hline
    Team & mAP (\%) \\
    \hline
    \hline    
    VIPL & \textbf{69.68} \\
    JDAI-Human & 64.40 \\
    NtechLab Team & 62.49 \\
    fourzerotwo & 61.17 \\
    ALFNet & 60.45 \\
    \hline
\end{tabular}
\end{table}

\subsection{Solution of First Place}

\begin{figure}[h]
    \centering
    \includegraphics[width=\linewidth]{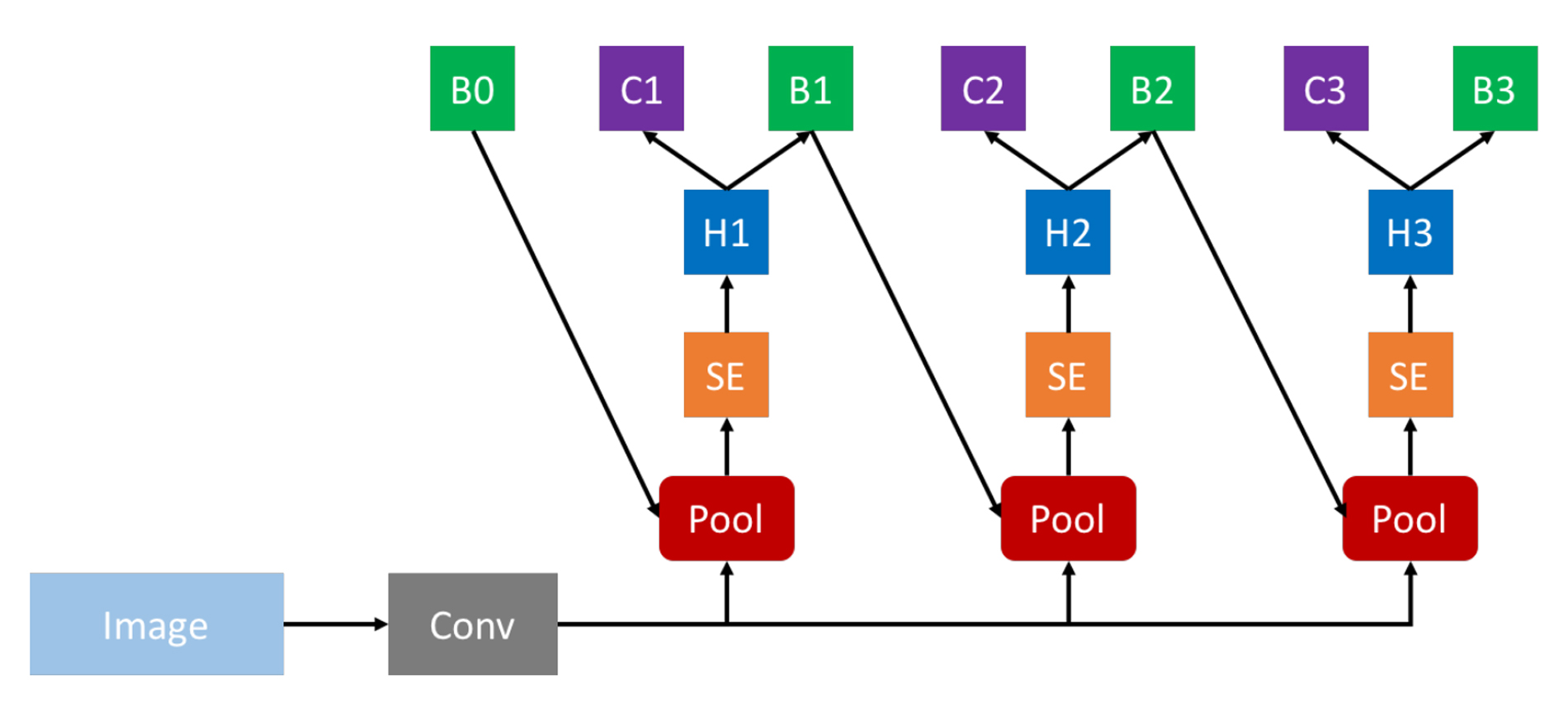}
    \caption{Framework of the WIDER Pedestrian Detection 1st-place solution.
    "Conv" is the backbone convolutions including deformable convolutions, "pool" RoI Align, "SE" channel-wise attention, "H" network head, "B" bounding box, and "C" classification. ”B0” is the proposals. }
    \label{fig:pedestrian-1st}
\end{figure}

The basic detection framework of the champion is Cascade R-CNN~\cite{cai2018cascade}. They also add some powerful structures to achieve better performances. The overall framework is shown in Fig.~\ref{fig:pedestrian-1st}.

\noindent\textbf{Framework}.
The backbone of the detection framework is FPN~\cite{lin2017feature} with deformable convolution~\cite{dai2017deformable} to extract features. Considering the large number of small-scale pedestrians, RoI-Align~\cite{he2017mask} replaced RoI-Pooling~\cite{ren2017faster} to align these small objects better. A channel-wise attention~\cite{chen2017sca} is added after pool5 to deal with the occlusion problem. 

\noindent\textbf{Implementation Details}.
(1) The training data are augmented through methods such as Gaussian blur and random cropping.
(2) Five models are ensembled: ResNet-50~\cite{he2016deep}, DenseNet-161~\cite{huang2017densely}, 197 SENet-154~\cite{hu2018squeeze} and two ResNext-101~\cite{xie2017aggregated} models. 
(3) Multi-scale input images are adopted at the testing stage. Specifically, four scales are adopted: [600, 1600],
[800, 2000], [1000, 2000] and [1200, 2400], the former number is the short size, and the latter one is the max size. The bounding boxes from different scales are merged and the final results come from the voting and soft-NMS of these merged bounding boxes.
(4) The classifier is trained with 3 classes, [background, person, cyclist]. The last two are merged as a person class when testing.

Table \ref{fig:pedestrian-1st_comp} shows the improvement of each component when compared with the baseline Res50-FPN. 

\begin{table}[h]
    \caption{Improvement of each component over the baseline Res50-FPN.}
    \centering
    \label{fig:pedestrian-1st_comp}
\begin{tabular}{c|c|c}
\hline
Method &Comments &Gain \\
\hline
Cascade RCNN & 3 stage [0.5, 0.6, 0.7] &3.8 \\
Deformable conv &-& 0.8 \\
Reweight Pool5 &-& 0.8 \\
Multi label &specify person and cyclist &0.4 \\
Augmentation &color and random crop &3.5 \\
Bn training &-& 1.3 \\
Multi-scale testing &4 scale with flip &2.9 \\
Ensemble &5 models& 2.2 \\
\hline
\end{tabular}
\end{table}

\subsection{Solution of Second Place}

\begin{figure}[h]
    \centering
    \includegraphics[width=\linewidth]{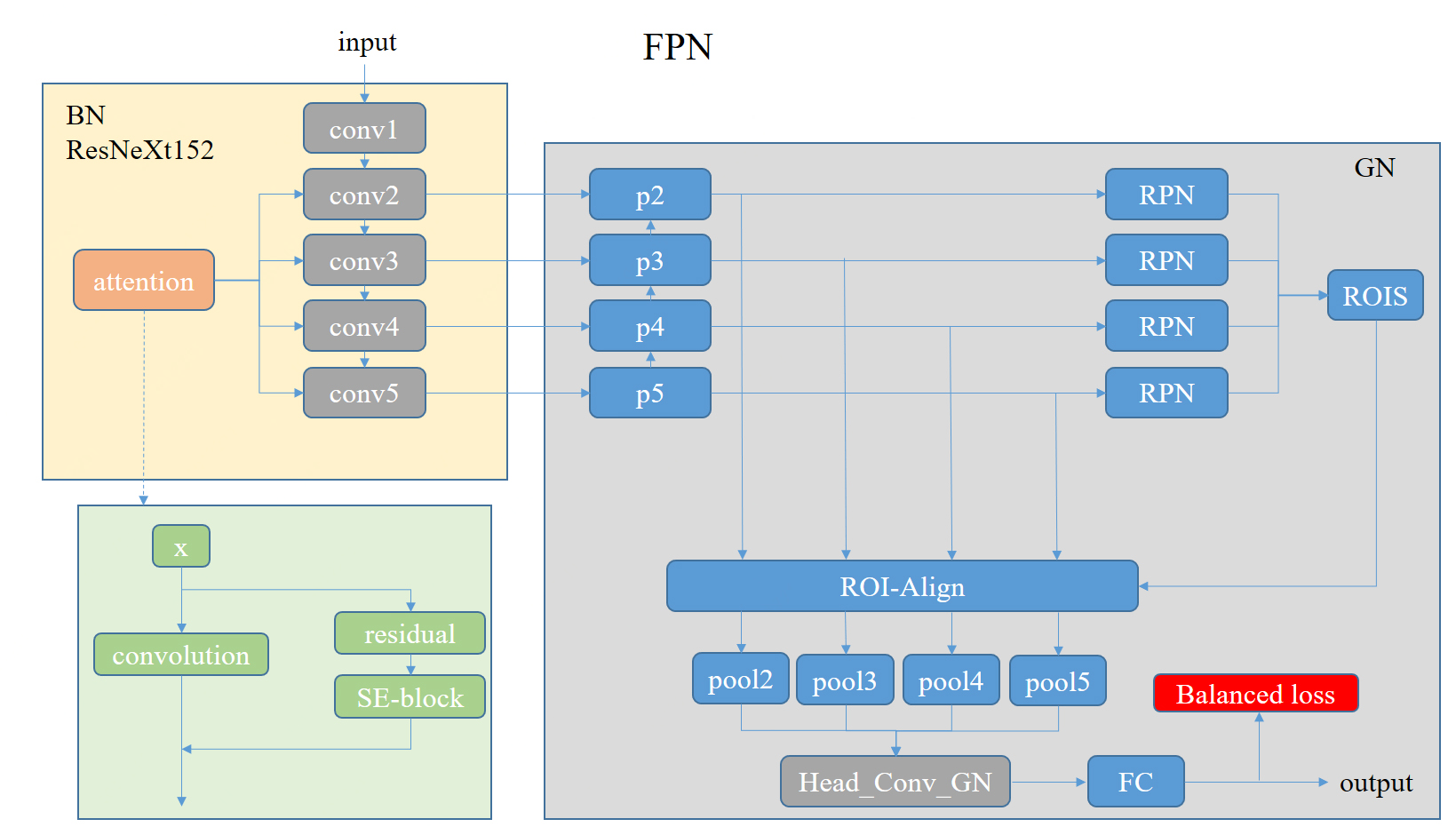}
    \caption{Framework of the WIDER Pedestrian Detection 2nd-place solution.}
    \label{fig:pedestrian-2nd}
\end{figure}

The second team uses FPN~\cite{lin2017feature} and Faster R-CNN~\cite{ren2017faster} as the basis of their detection framework. The overall framework is shown in Fig.~\ref{fig:pedestrian-2nd}. 

\noindent\textbf{Framework}.
    The backbone of the network is ResNext-152~\cite{xie2017aggregated}. Squeeze-and-Excitation blocks~\cite{hu2018squeeze} (SE-blocks) are added to every residual block as channel-wise attention. Batch Normalization~\cite{ioffe2015batch} is mixed with Group Normalization~\cite{wu2018group}. RoI Pooling~\cite{ren2017faster} is also replaced with RoI Align~\cite{he2017mask}.

\noindent\textbf{Implementation Details}.
    (1)The training data are augmented through horizontally-flip and multi-scale training. (2) The NMS is replaced with soft-NMS~\cite{bodla2017soft}. (3) The models are finetuned with stage-by-stage IoU threshold and focal loss function~\cite{lin2018focal} to improve the quality of the final bounding-boxes. 

\subsection{Solution of Third Place}
The team at the third place uses Cascade R-CNN~\cite{cai2018cascade} as the detection framework. The number of anchors is increased and data are also augmented with multi-scale training.

\subsection{Discussion}
The most challenging part of this track comes from a large amount of dense and small-scale pedestrians in the dataset. Almost all winners choose to design the detection framework on the basis of Cascade R-CNN to get better localizations of bounding boxes. To deal with the problem of a large number of small-scale pedestrians in the dataset, they choose to replace RoI Pooling with RoI Align. All these methods are quite effective and have achieved good performance. Most of the methods adopted here are mainly adopted from existing studies that found effective on object detection. In the next challenge, we expect new methods that are specifically designed for pedestrian detection.

%% file: sections/person_search.tex

\section{Person Search Track}
\label{sec:person-search}

\subsection{Task and Dataset}

To search for a person in a large-scale database with just a single portrait is a practical but challenging task. 
In the person search track of WIDER Challenge, given a portrait of a target cast and some candidates media (frames of a movie with person bounding boxes), one is asked to search for all the instances belonging to that cast.

Data used in this track is based on the \emph{Cast Search in Movies} (CSM) Dataset~\cite{huang2018person}. CSM  contains 127K tracklets of 1,218 cast from 192 movies.
In the WIDER Challenge, we model the person search problem as an image-based task by choosing one key frame of each tracklet as candidates.
Among the $192$ movies, $115$ are used for training, $19$ are used for validation and $58$ are for testing.
For each movie, the main cast (top $10$ in the cast list of IMDb) are collected as queries. The query profile comes from the homepage of the cast in IMDb or TMDb. The candidate's frames are extracted from the keyframes of the movie, in which the bounding boxes and identities of persons are manually annotated. A candidate is either annotated as one of the main casts or as "others". Here "others" means that a candidate does not belong to any of the main casts of that movie.
There are 1006 queries in training, 147 in validation and 373 in the test. The average number of candidates of each split are 690, 796 and 560 per movie.

\begin{figure}[h]
    \centering
    \includegraphics[width=\linewidth]{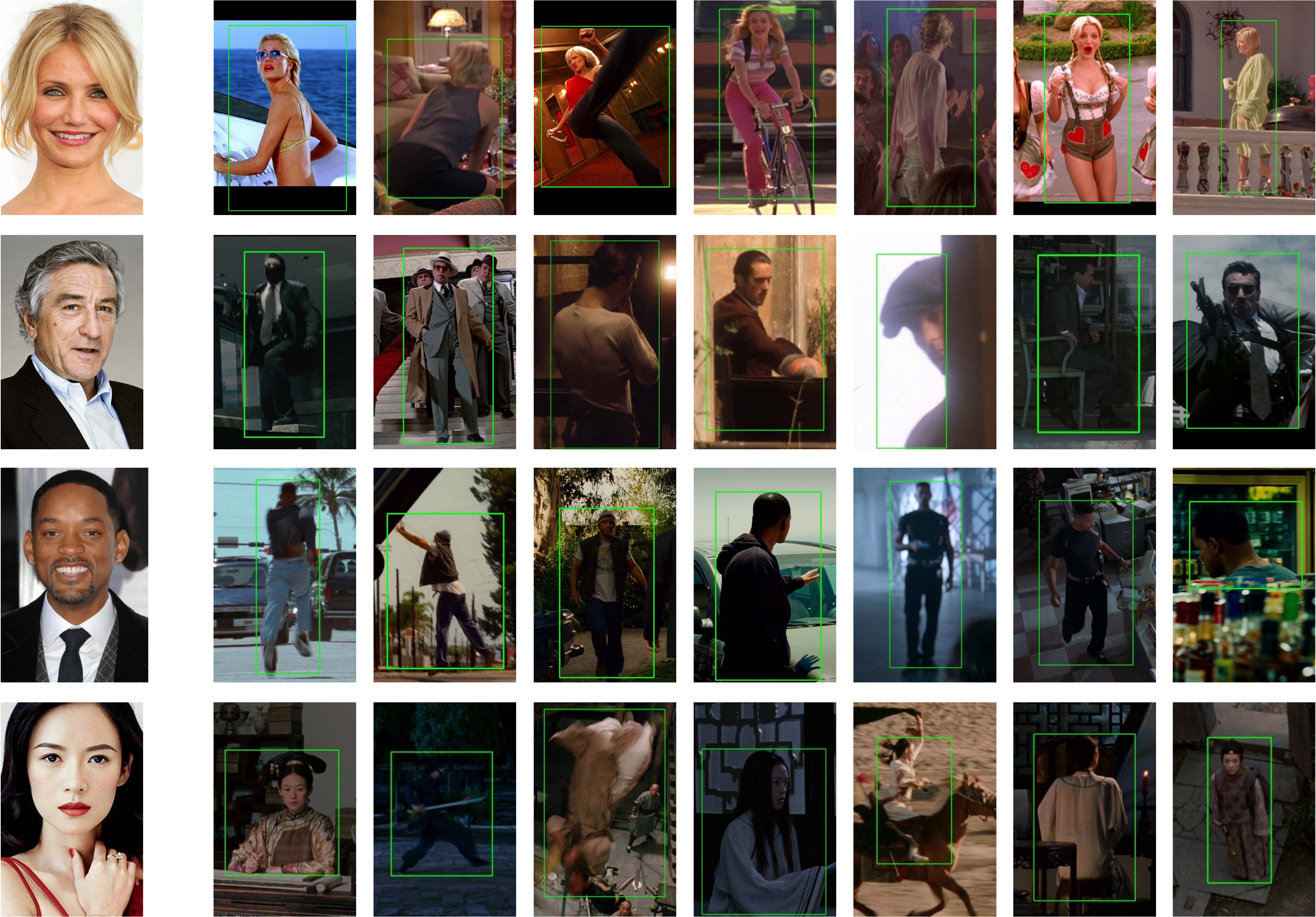}
    \caption{Examples of CSM Dataset used in Person Search Track. Images on the most left row are the portraits of the target cast, which are queries in this task. The other images in the same rows are their instances in the movies, which are the candidates in this track.}
    \label{fig:csm}
\end{figure}

\subsection{Evaluation Metric}

Following other retrieval tasks, here we use mean Average Precision (mAP) as our evaluation metric,
which can be formulated as Eq.~\ref{eq:map}.
\begin{equation}
\label{eq:map}
    mAP = \frac{1}{Q}\sum_{q=1}^{Q}\frac{1}{m_q}\sum_{k=1}^{n_q}P_q(k)rel_q(k)
\end{equation}

Here $Q$ is the number of query cast; $m_q$ is the number of candidates with the same identity to the query; $n_1$ is the number of all candidates in the movie; $P_q(k)$ is the precision at rank k for the q-th query; $rel_q(k)$ denotes the relevance of prediction k for the q-th query, it's 1 if the k-th prediction if correct and 0 otherwise.

\subsection{Results of the Challenge}

There are more than $100$ teams participate in the person search track of WIDER Challenge 2018. The results of the top5 teams are shown in Tab.~\ref{tab:person-result}.

\begin{table}[!h]
\renewcommand{\arraystretch}{1.3}
\caption{Results of the Top5 Teams in Person Search Track}
\label{tab:person-result}
\centering
\begin{tabular}{c|c|c}
    \hline
    Rank & Team & mAP (\%) \\
    \hline
    \hline    
    1 & Jiaoda Poets & \textbf{76.71} \\
    2 & SAT\_ICT & 74.66 \\
    3 & MCC\_USTC & 74.02 \\
    4 & TUM-MMK & 66.70 \\
    5 & ll490187880 & 66.38\\
    \hline
\end{tabular}
\end{table}

\subsection{Solution of First Place}

\begin{figure}[h]
    \centering
    \includegraphics[width=\linewidth]{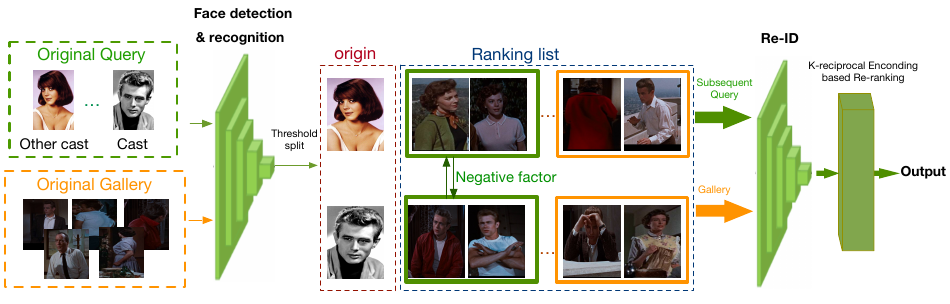}
    \caption{Framework of the 1st-place solution in Person Search.
        It contains face detection\&recognition, re-ranking based on face feature, person re-id, and re-ranking based on re-id feature.}
    \label{fig:person-1st}
\end{figure}

The winning team designs a cascaded model that utilizes both face and body features for person search.
The overall framework is shown in Fig.~\ref{fig:person-1st}.

\noindent\textbf{Pipeline}.
(1) An off-the-shelf face detector is used to detect faces on the dataset.
(2) Training a face verification model with an external dataset under cross entropy loss.
(3) Re-ranking the face matching result based on the Euclidean distance and Jaccard distance between the cast's face and candidates' faces.
(4) Setting a threshold score to split candidates to queries and galleries.
(5) Training a Re-ID model on the training set and then applying multi-query person Re-ID. Note that the matching results of other casts are taken as the negatives to reduce the amount of the galleries in Re-ID.
(6) Re-ranking the Re-ID matching result as what is done on face matching result.

\noindent\textbf{Implementation Details}.
(1) The face detector used here is MTCNN~\cite{zhang2016joint} trained on WIDER FACE~\cite{yang2016wider}.
(2) The face recognition model backbones include ResNet~\cite{he2016deep}, InceptionResNet-v2\cite{szegedy2017inception}, DenseNet~\cite{huang2017densely}, DPN and MobiletNet~\cite{howard2017mobilenets}.
(3) The Re-ID backbones include ResNet=50, ResNet-101, DenseNet-161 and DenseNet-201.

\subsection{Solution of Second Place}

\begin{figure}[h]
    \centering
    \includegraphics[width=\linewidth]{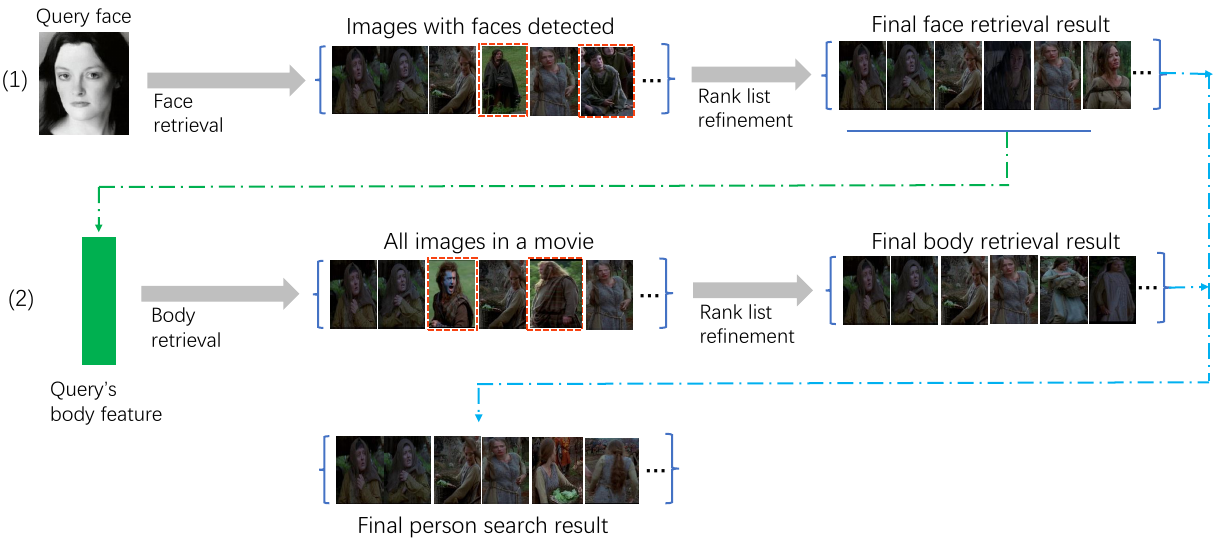}
    \caption{Framework of the 2nd-place solution in Person Search.
        The stage (1) is to retrieve faces, and the stage (2) is to retrieve bodies. The dashed green line indicates that the top-ranked gallery images are aggregated to find out the appearance of the query subject’s body. The dashed blue line represents that the two retrieval rank lists are fused as the ultimate results. Images in the dashed red box are imposters.}
    \label{fig:person-2nd}
\end{figure}

The solution is decomposed into two stages - the first stage is to retrieve
faces, and the second stage is to retrieve the bodies. Finally, the retrieval results
of the two stages are combined as the ranking result.
The overall framework is shown in Fig.~\ref{fig:person-2nd}.

\noindent\textbf{Pipeline}.
(1) Applying two algorithms to detect faces.
(2) Given a query face of a subject, those gallery images with faces are ranked by face retrieval.
(3) Rank list refinement.
(4) The top-ranked gallery images are adaptively aggregated to find out the appearance of the query subject’s body.
(5) All candidates are ranked by body retrieval.
(6) The retrieval results from face and body are fused in similarity score level

\noindent\textbf{Implementation Details}.
(1) Face Detection.
The face detector used here are PCN\cite{shi2018real} and MTCNN\cite{zhang2016joint}.
(2) Face Retrieval.
A second-order networks~\cite{lin2015bilinear,li2017second,chowdhury2016one} (ResNet-34 as backbone) trained on VGGFace2~\cite{cao2018vggface2} with softmax loss and ring loss~\cite{zheng2018ring} is used here.
It is ensembled with the provided ResNet-101 in facial cosine similarity score level with the same weights.
(3) Body Retrieval.
SE-ResNeXt50~\cite{xie2017aggregated,hu2017squeeze} with Residual Attention Network~\cite{wang2017residual} block is used here.
It is trained with both softmax loss and ring loss~\cite{zheng2018ring}.
(4) Re-Ranking.
Specifically, $d^{*}_{face}(p, g_i)$ on faces is calculated to
re-rank gallery images to get the re-rank distances as~\cite{zhong2017re}.
Then the re-rank distances are transfered into the similarities $s^{*}_{face}(p, g_i) = 1 - d^{*}_{face}(p, g_i)$.
Following that, the gallery images whose re-rank face similarity is greater
than $\theta$ are integrated by weighted average pooling to obtain the query subject’s
body feature $f_p$ as Eq.~\ref{eq:2nd-fp}.
\begin{equation}
\label{eq:2nd-fp}
    f_p = \sum_{i}^{n}f_{g_i} \times s^{*}_{face}(p, g_i), \text{~~s.t.~~}
     s^{*}_{face}(p, g_i) > \theta
\end{equation}
The re-rank body similarity $s^{*}_{body}(p, g_i)$ is calculated similarly, 
and the two similarities are fused by average.

\subsection{Solution of Third Place}

\begin{figure}[h]
    \centering
    \includegraphics[width=\linewidth]{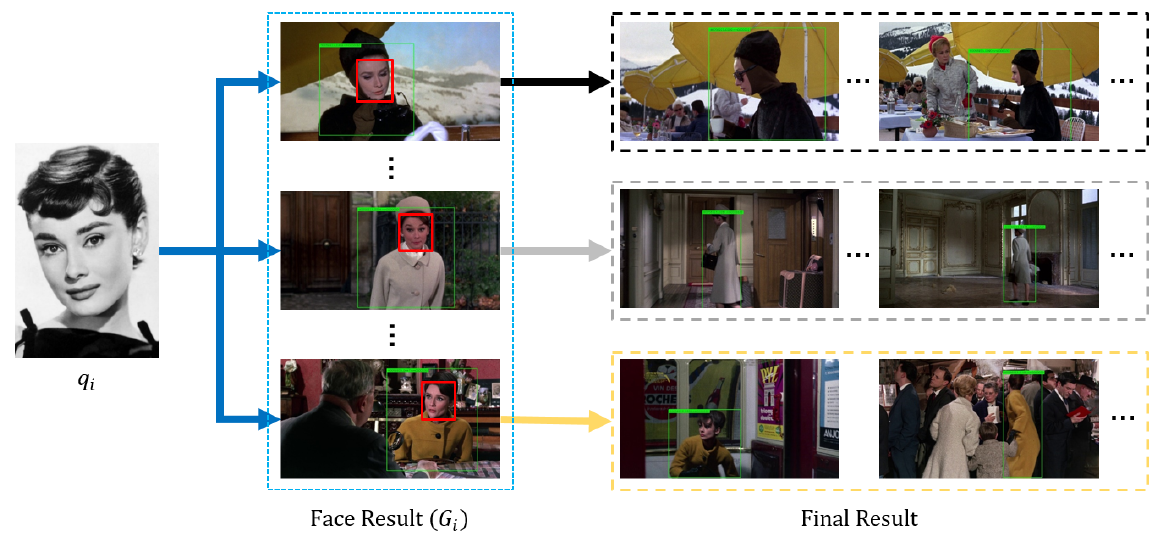}
    \caption{Framework of the 3rd-place solution in Person Search.
        Firstly, the preliminary collections of candidates $G_i$ corresponding to the $i$-th query $q_i$ are gotten . Then similar instances in all
        candidates are searched according to the ReID features.}
    \label{fig:person-3rd}
\end{figure}

A two-step framework is proposed to tackle this problem, which is shown in Fig.~\ref{fig:person-3rd}.
In the first step, the face in the query is used to search persons, whose faces can be detected, by face recognition.
So that a set of images relevant to the query can be obtained.
Then these images are further used to search again in all candidate images by person re-identification feature to get the final result.

\noindent\textbf{Pipeline}.
(1) Use two off-the-shelf face detectors for face detection.
(2) Get the preliminary collections of candidates corresponding to the query based on face recognition.
(3) Trained a person re-id model on WIDER-ReID (modified from CSM~\cite{huang2018person} under the person re-id setting).
(4) Retrieve based on the re-id feature.

\noindent\textbf{Implementation Details}.
(1) MTCNN~\cite{zhang2016joint} is used to detecting faces and ArcFace~\cite{deng2018arcface} is used for face recognition.
(2) The re-id model backbones include ResNet-101, DenseNet-121, SEResNet-101, and SEResNeXt-101.
(3) The final distance between query $q_i$ and candidate $g_j$ is computed as Eq.~\ref{eq:3rd-dist}, where $f$ is the re-id feature, $G_i$ is the preliminary collections of $g_i$ based on face recognition.
\begin{equation}
\label{eq:3rd-dist}
    d(q_i, g_j) = \min_{g_k \in G_i}||f_{g_k} - f_{q_i}||_2
\end{equation}

\subsection{Discussion}

The most challenging issue in this track is that the portrait contains only face while some of the candidates are without frontal faces.
Almost all participants choose to use a two-stage framework to tackle this problem.
The first stage performs face recognition and retrieves some confident instances based on the face feature.
The second stage uses body feature (re-id feature) for re-ranking,
so as to deal with all instances no matter with faces or not.
MTCNN~\cite{zhang2016joint} is widely used for face detection and many powerful networks structures like ResNet~\cite{he2016deep}, DenseNet~\cite{huang2017densely}, ResNeXt~\cite{Xie2016}, and SEResNet~\cite{hu2017squeeze}, are used as face and body recognition models.
The k-reciprocal encoding method~\cite{zhong2017re} is popular for re-ranking.

Although this two-stage framework yields good performances in this task,
A unified framework for extracting both face and body features is not yet proposed.
Also, semantically important information such as the scene and person relationship~\cite{li2016multi,huang2018unifying}, is rarely explored. 
We believe that there are still rooms for improvement for this task.